\documentclass[10pt,twocolumn,letterpaper]{article}

\usepackage{iccv}
\usepackage{times}
\usepackage{epsfig}
\usepackage{graphicx}
\usepackage{amsmath}
\usepackage{amssymb}
\usepackage{graphicx}
\usepackage{amsmath}
\usepackage{amssymb}
\usepackage{booktabs}

\usepackage{url}
\usepackage{epsfig}
\usepackage[accsupp]{axessibility}
\usepackage{helvet}   
\usepackage{courier}  
\usepackage{url}      
\usepackage{xcolor}
\usepackage{multirow}
\usepackage{subcaption}
\usepackage{algorithm}
\usepackage{algpseudocode}
\usepackage[export]{adjustbox}
\usepackage{color}
\usepackage{pifont}
\definecolor{mycolor}{RGB}{147,112,219}
\definecolor{yccoloer}{RGB}{52, 235, 150}
\usepackage{appendix}
\usepackage{amsmath}

\usepackage[pagebackref=true,breaklinks=true,letterpaper=true,colorlinks,bookmarks=false]{hyperref}

\iccvfinalcopy 

\def\iccvPaperID{1118} 

\begin{document}

\title{CLIP2NeRF-latent: View-Invariant Latent Diffusion via Contrastive Learning}
\title{NeRF-LaFusion: View-Invariant Latent Diffusion in NeRFs\\ for Realistic Text-to-3D Creation}
\title{CLIP2NeRFusion: View-Invariant NeRF Latent Diffusion with CLIP\\ for Domain-Specific Text-to-3D Rendering}
\title{Text2NeRFusion: View-Invariant Latent Diffusion\\ for Fast Text-to-3D Rendering}
\title{3D-LatentFusion: View-Invariant Latent Diffusion\\ for Fast Text-to-3D Rendering}
\title{3D-LatentFusion: Fast Text-to-3D Rendering\\ with View-Invariant Latent Diffusion}

\title{3D-CLFusion: Fast Text-to-3D Rendering with Contrastive Latent Diffusion\vspace{-5mm}}


\author{
Yu-Jhe Li$^{1*}$\qquad\qquad
Tao Xu$^{2*}$
\qquad\qquad Ji Hou$^{2*}$
\qquad\qquad Bichen Wu$^{2*}$
\qquad\qquad Xiaoliang Dai$^{2*}$\\ 
\qquad Albert Pumarola$^{2*}$
\qquad Peizhao Zhang$^{2*}$ \qquad Peter Vajda$^{2*}$ \qquad Kris Kitani$^{1*}$
\\
$^{1}$Carnegie Mellon University
\qquad\qquad
$^{2}$Meta\\
{\tt\small 
\{\url{yujheli},\url{kkitani}\}\url{@cs.cmu.edu}} \\
{\tt\small $^{*}$Work done in 2022 during Yu-Jhe Li's internship with Meta}
\vspace{-3mm}
}


\twocolumn[{%
\renewcommand\twocolumn[1][]{#1}%
\maketitle
\vspace{-10mm}

\begin{center}
    \centering
    \includegraphics[width=0.9\textwidth]{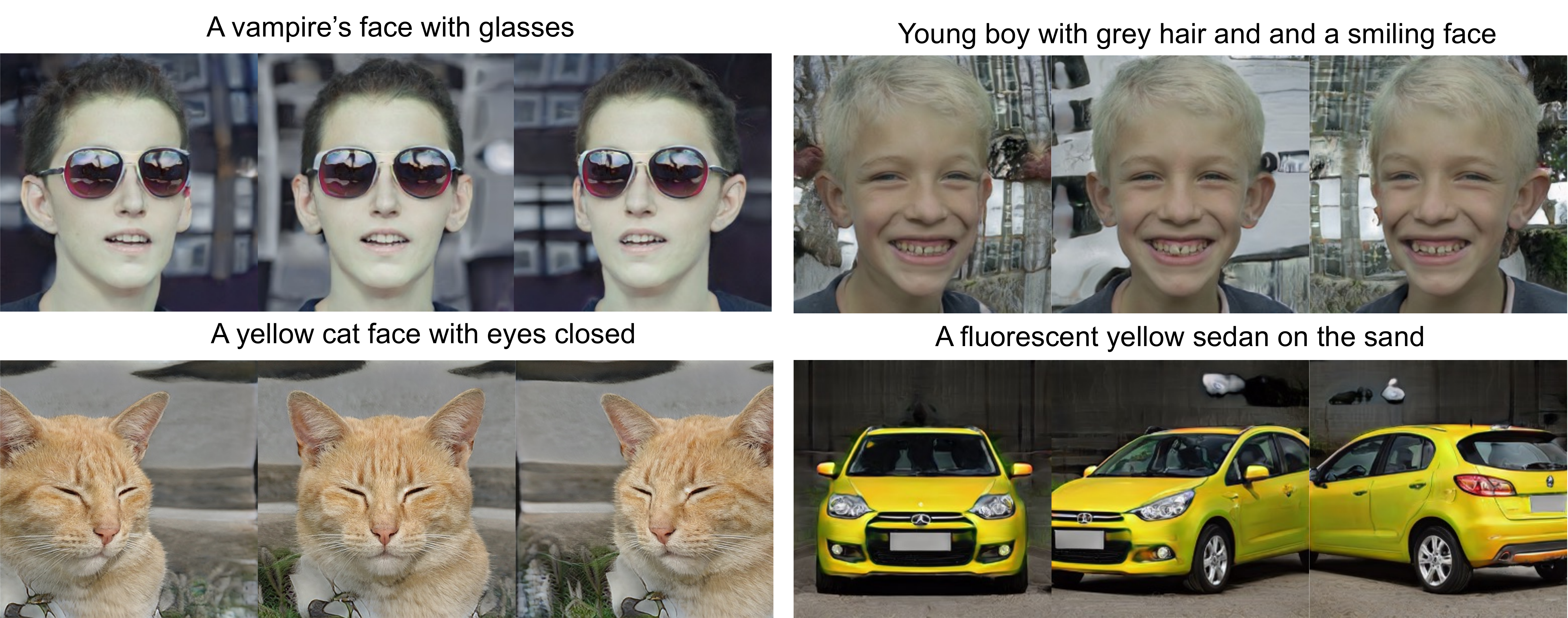}
    \vspace{-3mm}
    \captionof{figure}{Multi-view images generated from text prompts by 3D-CLFusion.}
    \label{fig:show}
    \vspace{-3mm}
\end{center}%
}]


\maketitle

\begin{abstract}
\vspace{-5mm}

We tackle the task of text-to-3D creation with pre-trained latent-based NeRFs (NeRFs that generate 3D objects given input latent code). Recent works such as DreamFusion and Magic3D have shown great success in generating 3D content using NeRFs and text prompts, but the current approach of optimizing a NeRF for every text prompt is 1) extremely time-consuming and 2) often leads to low-resolution outputs. To address these challenges, we propose a novel method named 3D-CLFusion which leverages the pre-trained latent-based NeRFs and performs fast 3D content creation in less than a minute. In particular, we introduce a latent diffusion prior network for learning the w latent from the input CLIP text/image embeddings. This pipeline allows us to produce the w latent without further optimization during inference and the pre-trained NeRF is able to perform multi-view high-resolution 3D synthesis based on the latent. We note that the novelty of our model lies in that we introduce contrastive learning during training the diffusion prior which enables the generation of the valid view-invariant latent code. We demonstrate through experiments the effectiveness of our proposed view-invariant diffusion process for fast text-to-3D creation, e.g., 100 times faster than DreamFusion. We note that our model is able to serve as the role of a plug-and-play tool for text-to-3D with pre-trained NeRFs.

\end{abstract}
\section{Introduction}


We aim the tackling the task of text-to-3D domain-specific content creation. 3D content can be represented with neural radiance field (NeRF)~\cite{mildenhall2020nerf} in a photorealistic way. Currently, text-to-3D with NeRFs have been explored in DreamField~\cite{jain2022zero}, DreamFusion~\cite{poole2022dreamfusion}, or Magic3D~\cite{lin2022magic3d}. By leveraging the pre-trained models from CLIP~\cite{radford2021learning} or diffusion priors~\cite{saharia2022photorealistic} as the learning objective, these works are capable of producing 3D content given the input text prompt through training a NeRF from scratch. However, training one individual model for each text prompt would cause the first issue: slow inference (\textit{i.e., around one hour}) for the above models. Due to no real image guidance during updating, the model would lead to the second issue: low-resolution multi-view rendering.

\begin{figure}[t!]
  \centering
  \includegraphics[width=\linewidth]{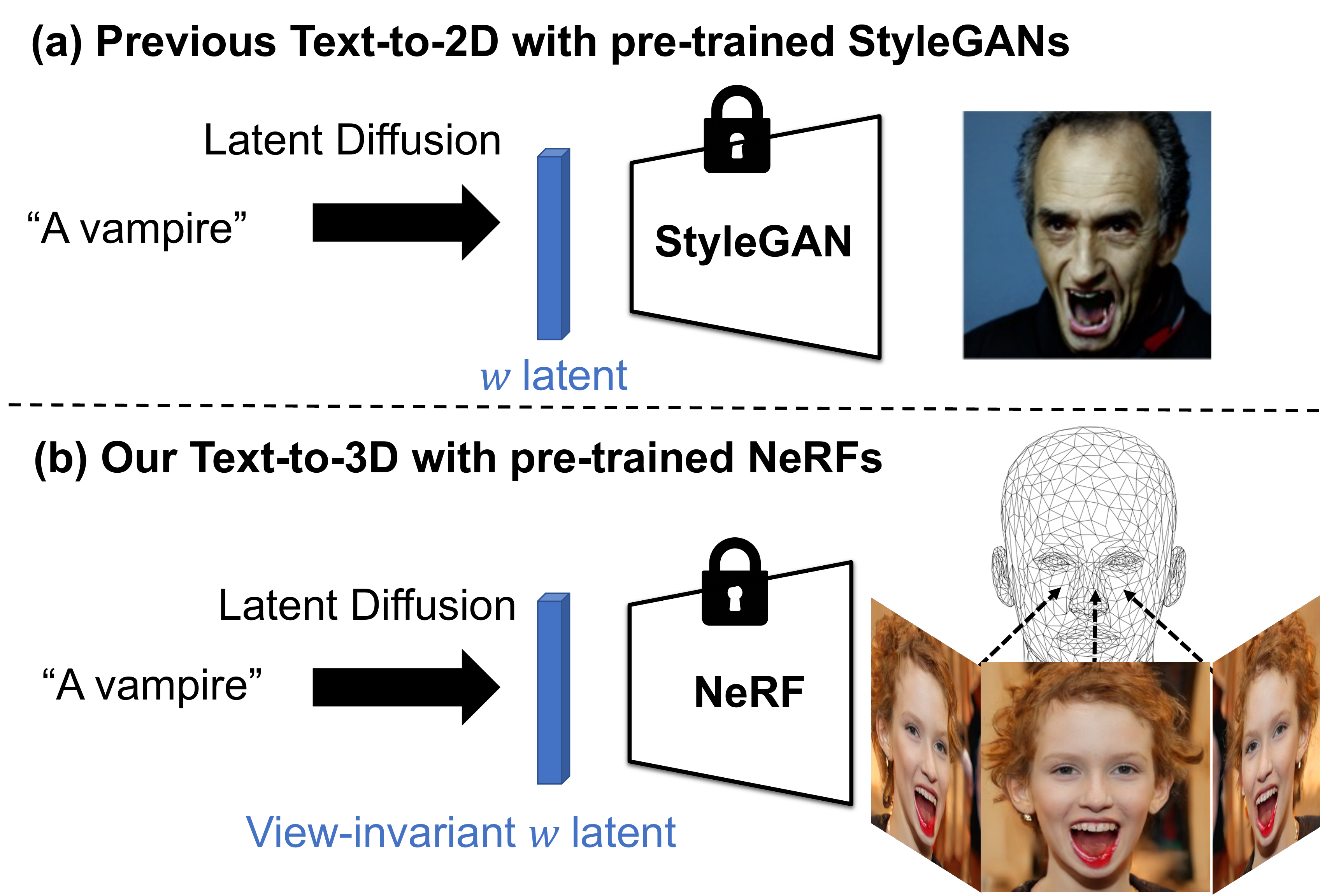}
  \caption{\textbf{Comparison of text to content creation with pre-trained GANs}. Compared with existing effective text-to-2D with StyleGANs~\cite{pinkney2022clip2latent}, it is more challenging to perform text-to-3D with latent-based NeRFs via diffusion process since the denoised latent is assumed to be view-invariant. }
  \label{fig:teasor}
  \vspace{-4mm}
\end{figure}

Recently, 3D latent-based models using radiance fields (\textit{i.e.}, NeRFs~\cite{mildenhall2020nerf}), such as EG3D~\cite{chan2022efficient} or StyleNeRF~\cite{gu2021stylenerf}, have been proposed for unsupervised generation of multi-view consistent images. Similar to StyleGANs~\cite{karras2019style,Karras2019stylegan2}, these NeRFs learn a controllable $w \in \mathcal{W}$ space and enable explicit 3D camera control, using only single-view training images. 
To achieve fast text-to-3D creation, one straightforward way is to produce the $w$ latent from the input text prompt without further updating the NeRF model. This idea has been explored in clip2latent~\cite{pinkney2022clip2latent} for text-to-2D creation, which takes less than one minute to generate high-resolution realistic images from the input text prompt with a trained diffusion prior. However, there are two main challenges for directly applying the clip2latent on 3D latent-based NeRF. First, the latent space $w \in \mathcal{W}$ is assumed to be view-invariant for 3D NeRFs which is different from 2D StyleGANs. That is, if only single-view prompts (image CLIP embeddings) are used to train the diffusion prior, the produced latent code may be only valid when generating images of the same view (camera pose) where this issue is also noted in NeRF-3DE~\cite{li20223d}. To address this issue, we can use multi-view CLIP prompts from the same latent $w$ to train the diffusion prior. However, this leads to our second challenge: how to ensure the diffusion process to produce view-invariant latent from multi-view CLIP embeddings, as shown in Figure~\ref{fig:teasor}.

To properly achieve fast and realistic text-to-3D creation with pre-trained NeRFs, we propose a framework named 3D-CLFusion to produce the view-invariant latent code from an input text prompt. Our 3D-CLFusion is composed of a diffusion prior network, a pre-trained clip model, and a pre-trained generator (\textit{i.e., NeRF}). First, in order to produce the latent code $w$ for the generator to render from the input text prompt, we introduce the diffusion prior which takes the CLIP text embedding as a condition and produces $w$ latent in each diffusion step. Since we do not have labeled text embeddings, we leverage the image CLIP embeddings for training the diffusion prior. This training strategy is inspired by clip2latent~\cite{pinkney2022clip2latent} where we believe CLIP text and image embeddings share the closed latent space. Second, in order to learn the view-invariant latent in $\mathcal{W}$ space, we leverage the multi-view images generated by the model itself and contrastive learning to ensure the produced latent code in $\mathcal{W}$ are the same given different CLIP embeddings from a different view.
Later in the experiments, we will show the significance of our introduced contrastive learning in the diffusion process. We have demonstrated the effectiveness of the proposed framework for the fast text-to-3D using StyleNeRF~\cite{gu2021stylenerf} and EG3D~\cite{chan2022efficient} as the pre-trained generators in Figure~\ref{fig:show} and the experiments. Compared with DreamFusion~\cite{poole2022dreamfusion} and Magic3D~\cite{lin2022magic3d} which take around one hour for inference with NeRF, our model only takes less than 30 seconds for each 3D object. The contributions of this paper are summarized as follows:

\vspace{-3mm}
\begin{itemize}
\itemsep -1mm
\item We demonstrate the challenges of the task text-to-3D directly using latent-based NeRFs and the limitations of the current models for this task.
\item We propose a framework named 3D-CLFusion, which aims to produce view-invariant latent for rendering 3D objects with NeRFs from the input text prompt.
\item Compared with the existing models leveraging diffusion priors for text-to-2D with pre-trained StyleGANs, our designed model achieves more effective text-to-3D with pre-trained NeRFs.
\item Though the 3D object created by our model is limited to the class of pre-trained model, the inference time is at least 100 times faster than the existing text-to-3D creation from neural rendering.

\end{itemize}

\section{Related Works}



\paragraph{Text-to-image generation.}

There are several text-to-image generative models that have been proposed in recent months. The majority of these models are trained using large amounts of text-image paired data where image generation is conditioned on the text input. To achieve high-quality and accurate image generation from text, several of the existing approaches make use of pre-trained CLIP to produce text embedding. One line of research is to leverage diffusion models in \cite{ramesh2022hierarchical,rombach2022high,saharia2022photorealistic}, where the models directly learn the mapping between text and image with diffusion priors (\textit{i.e.}, Dalle 2~\cite{ramesh2022hierarchical} and Glide\cite{nichol2021glide}) or sample from a low-resolution latent space and decode the latent into high-resolution images (\textit{i.e.}, Stable Diffusion~\cite{rombach2022high}). Another line of work~\cite{patashnik2021styleclip,gal2022stylegan,abdal2022clip2stylegan,kocasari2022stylemc} is to perform text-guided generation or editing relying on CLIP and the pre-trained StyleGANs~\cite{karras2019style,Karras2019stylegan2}. StyleGANs have achieved high image quality and support different levels of semantic manipulation in the $w \in \mathcal{W}$ latent space.
Recently, clip2latent~\cite{pinkney2022clip2latent} has been proposed to produce the $w$ latent in StyleGAN from the input text prompt with the diffusion model.
However, most of these works focus on the rendering of 2D images
and are not capable of manipulating camera poses easily as NeRF~\cite{mildenhall2020nerf}.

\paragraph{3D image synthesis with NeRFs.}

Methods built on implicit
functions, e.g., NeRF~\cite{mildenhall2020nerf}, have been proposed in \cite{chan2021pi,schwarz2020graf,pan2021shading,niemeyer2021giraffe}.
To generate high-resolution images conditioned on the input style latent code, EG3D~\cite{chan2022efficient}, StyleNeRF~\cite{gu2021stylenerf}, VolumeGAN~\cite{xu20223d}, StyleSDF~\cite{or2022stylesdf}, and GMPI~\cite{zhao2022generative} have been developed. In addition, some works such as Sofgan~\cite{chen2022sofgan} and Sem2NeRF~\cite{chen2022sem2nerf} are able to perform multi-view synthesis with NeRF by taking into multi-view or single-view semantic masks.
However, most of these works are not capable of generating 3D objects given purely input text. We will demonstrate the effectiveness of our proposed diffusion prior to serving as a text-to-3D plug-and-play tool into EG3D~\cite{chan2022efficient} and StyleNeRF~\cite{gu2021stylenerf} in this work.

\paragraph{Text-to-3D generation with NeRFs.}

In recent years, several models~\cite{jain2022zero,poole2022dreamfusion,lin2022magic3d} for the task of text-to-3D generation have been proposed using NeRFs~\cite{mildenhall2020nerf}. Dream Fields~\cite{jain2022zero} leverage the pre-trained image-text encoder (\textit{i.e., CLIP}~\cite{radford2021learning}) as the image guidance to optimize the neural implicit scene representations (\textit{i.e., NeRFs}) through online training. Since the pre-trained CLIP models may not be effective image-level generation objectives, some works such as DreamFusion~\cite{poole2022dreamfusion} and Magic3D~\cite{lin2022magic3d} turn to train the NeRF using pre-trained diffusion prior~\cite{saharia2022photorealistic} instead of CLIP.
Though DreamFusion~\cite{poole2022dreamfusion} and Magic3D~\cite{lin2022magic3d} are capable of performing satisfactory 3D content creation and rendering with an open-vocabulary input text prompt, the inference takes 1.5 hours and 40 minutes for each model to train a NeRF from scratch. In this work, we aim to resolve the issue by leveraging the pre-trained NeRFs and diffusion prior, and producing the latent in less than a minute.


\section{The Proposed Method}
\label{sec:method}

\subsection{Problem Formulation and Overview}

Given the input text prompt, our goal is to generate the conditioned multi-view images with the pre-trained latent-based NeRF generator as our text-to-3D task. Specifically, given the input text embedding with CLIP, denoted as $e_t$, we aim to produce the corresponding $w$ latent\footnote{latent $w \in \mathcal{W}$~\cite{shen2020interpreting} or the extended latent $w \in \mathcal{W}+$~\cite{Karras2019stylegan2}} as output. The pre-trained NeRF generator denoted as $G$, is able to synthesize images $x$ given different camera poses $p$: $x = G(w,p)$. 

\begin{figure*}[t!]
  \centering
  \includegraphics[width=\linewidth]{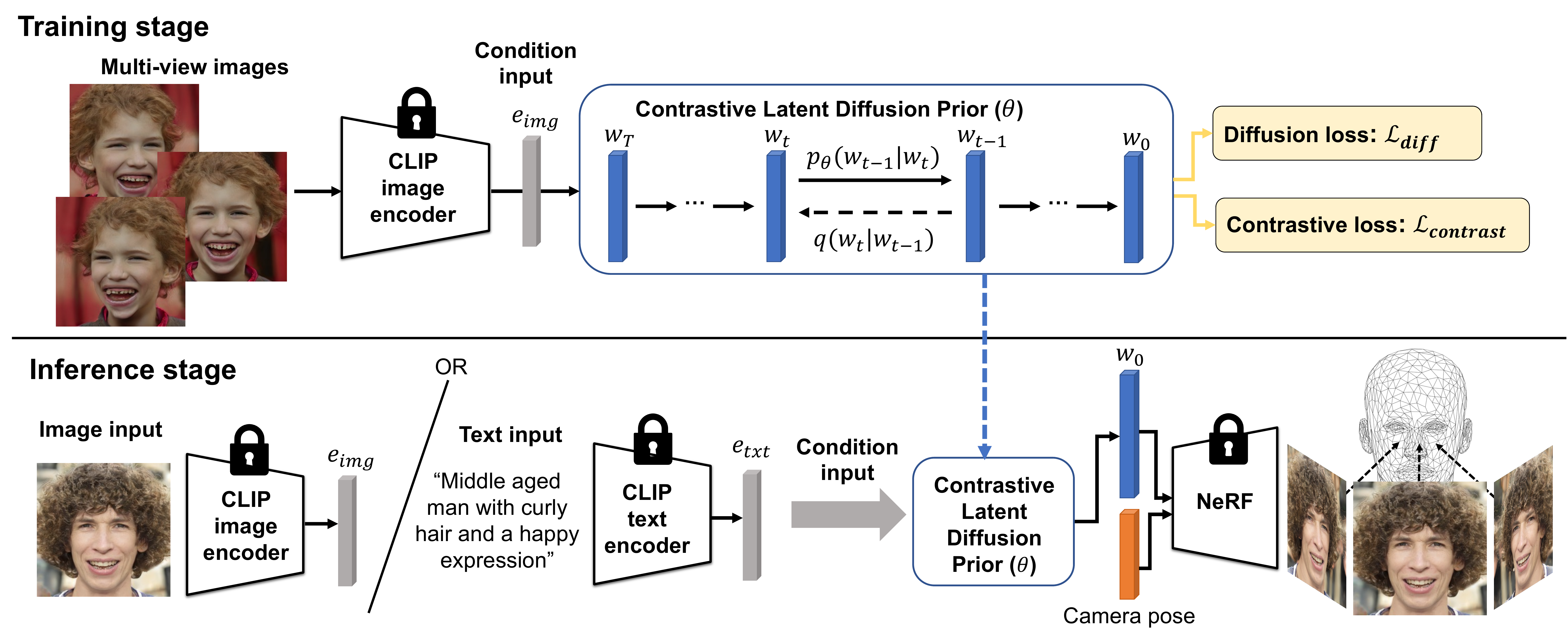}
  \vspace{-6mm}
  \caption{\textbf{Overview of our proposed 3D-CLFusion.} It consists of three modules: CLIP text/image encoders, contrastive latent diffusion prior, and pre-trained latent-based NeRF. More details can be referred to the section~\ref{sec:method}.}
  \label{fig:model}
\end{figure*}

In order to achieve text-to-3D creation by producing the $w$ latent for fast NeRF rendering, we propose a generative framework named 3D-CLFusion, which is presented in Figure~\ref{fig:model}. First, in order to produce the latent $w$ for the generator to render from the input text prompt, we introduce the diffusion prior network ($f_{\theta}$ where $\theta$ denotes the network parameters) which is able to take the CLIP text embedding $e_{txt}$ as an input condition and produces $w_0$ latent in the last diffusion step. More details of the diffusion/denoising process will be presented later. Since we do not have labeled text embeddings for the ground-truth latent $w_0$, we leverage the image CLIP embeddings for training the diffusion prior. We assume CLIP text and image embeddings share the closed latent space. Second, in order to learn the view-invariant latent $w$ in $\mathcal{W}$ space, we leverage the multi-view CLIP image embeddings from the images generated by the model and apply the contrastive loss to ensure the produced latent in $\mathcal{W}$ are view-invariant given different CLIP embeddings from a different view. After the training stage, we can sample text CLIP embedding from a given text input prompt and generate the corresponding 3D multi-view images with the pre-trained NeRF.


\subsection{Pre-trained Latent-based NeRF}
\paragraph{Latent-based Neural Implicit Representation.} Following StyleGANs~\cite{karras2019style,Karras2019stylegan2}, Latent-basde NeRFs~\cite{gu2021stylenerf} also introduce the mapping network $f$ which maps noise vectors from a spherical Gaussian space $\mathcal{Z}$ to the style space $\mathcal{W}$. $f$ consists of several MLP layers and the input style vector $w \in \mathcal{W}$ can be derived by $w = f(z), z \in \mathcal{Z}$. Following the neural rendering mechanism in NeRF~\cite{mildenhall2020nerf}, all of our pre-trained latent-based NeRF also takes the position $u\in \mathbb{R}^3$ and viewing direction $d \in \mathbb{S}^2$ as inputs, and predicts the density $\sigma(u)\in \mathbb{R}$ and view-dependent color $c(u,d) \in \mathbb{R}^3$.


\paragraph{Volume Rendering with Radiance Fields.}
Once we have the color and density for each coordinate and view direction, we render the color $C(r)$ for each pixel along that camera ray $r(t)=o+td$ passing through the camera center $o$ with volume rendering~\cite{kajiya1984ray}:
\begin{equation}\small
\begin{split}
C_{w}(r) = \int_{t_n}^{t_f} T(t)\sigma_{w}(r(t))c_{w}(r(t),d)dt,\\
\text{where}\quad T(t) = \exp(-\int_{t_n}^t \sigma_{w}(r(s))ds).
\end{split}
\end{equation}
The function $T(t)$ denotes the accumulated transmittance along the ray from $t_n$ to $t$. In practice, the continuous integration is discretized by accumulating sampled points along the ray. More details can be obtained in \cite{mildenhall2020nerf,gu2021stylenerf,chan2022efficient}.

\subsection{Latent Diffusion with CLIP embedding}
In order to produce the corresponding $w$ from the input text prompt $e_{txt}$ or image prompt $e_{img}$, we choose to employ a latent diffusion process to learn the mapping given the success of latent diffusion models in \cite{rombach2022high,pinkney2022clip2latent,ramesh2022hierarchical}. Now we present the overview of the latent diffusion model which contains: forward (diffusion) and backward (denoising) processing.

As stated in DDPM~\cite{ho2020denoising} and Latent Diffusion~\cite{rombach2022high}, we can formulate the diffusion process of $w$ latent diffusion for our task as:
\vspace{-3mm}
\begin{equation}\label{eq:forward}
\begin{split}
q(w_{1:T}|w_0) & =  \prod_{t=1}^{T} q(w_t|w_{t-1}),\\
 q(w_t|w_{t-1}) & = \mathcal{N}(w_t; \sqrt{1-\beta_t}w_{t-1},\beta_t \mathbf{I}),
\end{split}
\end{equation}
where $w_1$ ... $w_T$ are the latent of the same dimensionality of $w_0$ and $\beta_1<$ ... $<\beta_T$ are variance schedule. The reverse process can also be formulated as:
\begin{equation}\label{eq:backward}
\begin{split}
p_{\theta}(w_{0:T}) & = p(x_T) \prod_{t=1}^{T} p_{\theta}(w_{t-1}|w_{t}),\\
 p_{\theta}(w_{t-1}|w_{t}) & = \mathcal{N}(w_{t-1}; \mu_{\theta}(w_t,t),\Sigma_{\theta} (w_t,t)),
\end{split}
\end{equation}
where $\theta$ is the learnable parameters. If we set $\Sigma_{\theta} (x_t,t) = \sigma_t^2 \mathbf{I}$ as fixed variance, we only need to learn $\mu_{\theta}(w_t,t)$. Since we can denote $\alpha_t = 1-\beta_t$ and $\Bar{\alpha}_t = \prod_{s-1}^t \alpha_s$, we can estimate the forward process posteriors conditioned on $w_0$ as:
\begin{equation}\label{eq:posterior}
\begin{split}
 q(w_{t-1}|w_{t},w_0) = \mathcal{N}(w_{t-1}; \Tilde{\mu}_t (w_t,w_0), \Tilde{\beta}_t \mathbf{I}),\\
 \Tilde{\mu}_t (w_t,w_0) = \frac{\sqrt{\Bar{\alpha}_{t-1}}\beta_t}{1 - \Bar{\alpha}_{t}} w_0 + \frac{\sqrt{\alpha_t}(1 - \Bar{\alpha}_{t-1})}{1 - \Bar{\alpha}_{t}}w_t,\\
\Tilde{\beta}_t = \frac{1 - \Bar{\alpha}_{t-1}}{1 - \Bar{\alpha}_{t}}\beta_t.
\end{split}
\end{equation}
As stated in DDPM~\cite{ho2020denoising}, we can choose to learn the $\mu_{\theta}(w_t,t)$ in Equ.~\ref{eq:backward} and reparameterize it as:
\begin{equation}\label{eq:mu1}
\begin{split}
\mu_{\theta}(w_t,t) & = \Tilde{\mu}_t (w_t,w_0) = \frac{1}{\sqrt{\alpha_t}} (w_t - \frac{\beta_t}{\sqrt{1-\Bar{\alpha_t}}} \epsilon_\theta (w_t,t,e)),\\
\text{and}\quad w_0 & = \frac{1}{\sqrt{\alpha_t}} (w_t- \sqrt{1-\Bar{\alpha_t}}\epsilon_\theta (w_t,t,e))
\end{split}
\end{equation}
where $\epsilon_\theta$ is a function approximator intended to predict the noise $\epsilon$ from $w_t$, timestep embeddings $t$, and the CLIP embeddings $e_{txt}$ or $e_{img}$. Or we can reparameterize $\mu_{\theta}(w_t,t)$ as:
\begin{equation}
\begin{split}
& \mu_{\theta}(w_t,t) = \Tilde{\mu}_t (w_t,w_0) \\ & = \frac{\sqrt{\Bar{\alpha}_{t-1}}\beta_t}{1 - \Bar{\alpha}_{t}} f_\theta (w_t,t,e) + \frac{\sqrt{\alpha_t}(1 - \Bar{\alpha}_{t-1})}{1 - \Bar{\alpha}_{t}}w_t,
\end{split}
\label{eq:mu2}
\end{equation}
where $f_\theta$ is a function approximator intended to predict the $w_0$ from $w_t$, timestep embeddings $t$, and the CLIP embeddings $e_{txt}$ or $e_{img}$. Later we will show learning $f_\theta$ in Equ.~\ref{eq:mu2} is preferable for contrastive learning compared with learning $\epsilon_\theta$ in Equ.~\ref{eq:mu1}. The diffusion prior loss can be defined as:
\begin{equation}\label{eq:loss_diff}
\begin{split}
\mathcal{L}_{diff} = \mathbb{E}[\|\epsilon - \epsilon_\theta (w_t,t,e))\|^2]\\ \text{or} \quad \mathcal{L}_{diff} = \mathbb{E}[\|w_0 - f_\theta (w_t,t,e)\|^2]
\end{split}
\end{equation}

\subsection{View-invariant Latent Diffusion}

For 3D latent-based NeRF diffusion, the latent $w$ not only needs to match the input CLIP embedding $e$ but is also assumed to be view-invariant. As stated in NeRF-3DE~\cite{li20223d}, only the latent $w$ inside the valid manifold in the latent space $\mathcal{W}$ for latent-based NeRF is capable to produce reasonable view-invariant 3D objects, while the latent outside the manifold would lead to severe 3D distortion.

Since the CLIP image embeddings generated from the same $w$ with different camera poses $p_i$, $e_{img}^i$ representing CLIP image embeddings from different views are also different. Hence, the produced $w_t$ using Equ.~\ref{eq:backward} would not be view-invariant. We need to ensure the latent produce from the diffusion prior in each step are view-invariant as:
\begin{equation}\label{eq:ideal}
\begin{split}
f_\theta(w_t,t,E_{clip}^{img}(x_i) & = f_\theta(w_t,t,E_{clip}^{img}(x_j)\\
x_i & = G(w_0, p_i), i\neq j
\end{split}
\end{equation}
where $p_i$, $i=0,1,..,n$ represent camera poses. 
Minimizing the objective allows the model to learn the view-invariant latent code $\hat{w}$ since it maps multi-view embeddings $e_i = E_{clip}^{img}(x_i)$ (controlled by the pose $p_i$) to the same latent code $w$ for each set of the training sample. During inference, a single-view CLIP embedding (either text or image embeddings) is mapped to the latent code $\hat{w}$ which can produce multi-view images of the same identity by changing the poses.

To ensure this, we can use contrastive learning to train the diffusion prior by applying constraints on $\Tilde{w}_0 = f_{\theta} (w_t, t, e)$ in Equ.~\ref{eq:mu2}. 
Specifically, we perform contrastive learning with L2 loss $\mathcal{L}_{2}$ and triplet loss $\mathcal{L}_{tri}$ on the produced $w_0$ in each diffusion step. We can formulate $\mathcal{L}_{2}$ loss as:
\begin{equation}\label{eq:l2}
\begin{split}
\mathcal{L}_{2} & = \|f_{\theta} (w_t, t, e^i)- f_{\theta} (w_t, t, e^j)\|^2,
\end{split}
\end{equation}
where $e_i$ and $e_j$ are produced by the same ground-truth $w$. To maximize the inter-class discrepancy while minimizing intra-class distinctness, we introduce $\mathcal{L}_{tri}$. Specifically, for each input image embedding $e$, we sample a positive image embedding $e_\mathrm{pos}$ with the same identity label and a negative image $e_\mathrm{neg}$ with different identity labels to form a triplet tuple. Then, the following equations compute the distances between $e$ and $e_\mathrm{pos}$/$e_\mathrm{neg}$:

\begin{equation}
  \begin{aligned}
  d_\mathrm{pos} = \| f_{\theta} (w_t, t, e) - f_{\theta} (w_t, t, e^{pos})\|_2, \\
  d_\mathrm{neg} = \|f_{\theta} (w_t, t, e) - f_{\theta} (w_t, t, e^{neg})\|_2,
  \end{aligned}
  \label{eq:d-pos}
\end{equation}
With the above definitions, we have the triplet loss $\mathcal{L}_{tri}$ defined as:
\begin{equation}
  \begin{aligned}
  \mathcal{L}_{tri} &~ 
  = \max(0, m + d_\mathrm{pos} - d_\mathrm{neg}),
  \end{aligned}
  \label{eq:tri}
\end{equation}
where $m > 0$ is the margin used to define the distance difference between the positive image pair $d_\mathrm{pos}$ and the negative image pair $d_\mathrm{neg}$. The contrastive loss can be summed up as:
\begin{equation}
  \begin{aligned}
  \mathcal{L}_{contrast} = \mathcal{L}_{2} + \mathcal{L}_{tri}
  \end{aligned}
  \label{eq:feat}
\end{equation}

We would like to note that, the contrastive loss can still be applied using Equ.~\ref{eq:mu1} on the predicted $\Tilde{w}_0 = \frac{1}{\sqrt{\alpha_t}} (w_t, \sqrt{1-\Bar{\alpha}_t}\epsilon_\theta (w_t,t,e))$. However, applying constraints on this $\Tilde{w}_0$ would not be stable since it depends on both predicted $\epsilon$ and $w_t$ in each step, where $w_t$ is varying and sampled from Gaussian in each step.

The total loss $\mathcal{L}$ for training our proposed NeRF-3DE is summarized as follows:
\begin{equation}
  \begin{split}
  \mathcal{L}_{total} & = \lambda_{diff}\cdot\mathcal{L}_{diff} + \lambda_{contrast}\cdot\mathcal{L}_{contrast},
  \end{split}
  \label{eq:fullobj}
\end{equation}
where $\lambda_{diff}$ and $\lambda_{contrast}$ are the hyper-parameters used to control the weighting of the corresponding losses.

\section{Experiment}
\label{sec:EXP}

\begin{figure*}[t!]
  \centering
  \includegraphics[width=\linewidth]{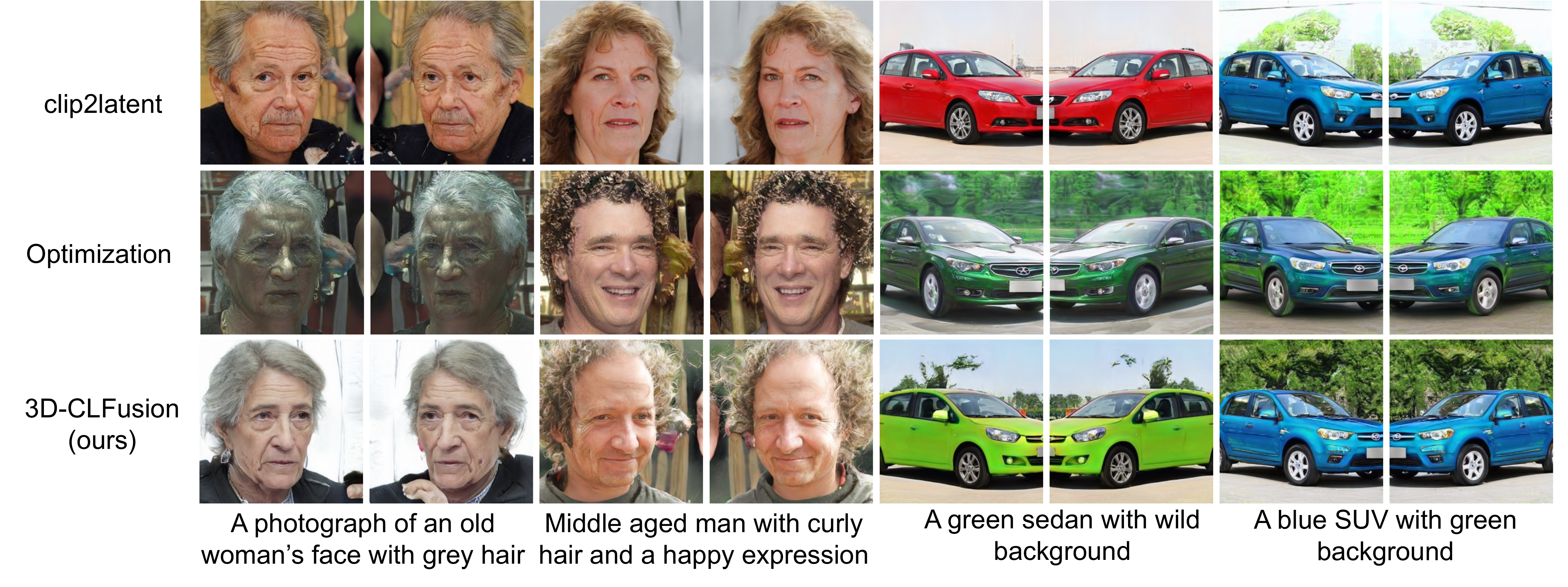}
  \vspace{-7mm}
  \caption{\textbf{Qualitative comparisons on text-to-3D with pre-trained latent-based NeRF: StyleNeRF~\cite{gu2021stylenerf}.} All of the output images are rendered using the same camera poses and the checkpoints from FFHQ and CompCars datasets.}
  \label{fig:stylenerf}
\end{figure*}

\begin{figure*}[t!]
  \centering
  \includegraphics[width=\linewidth]{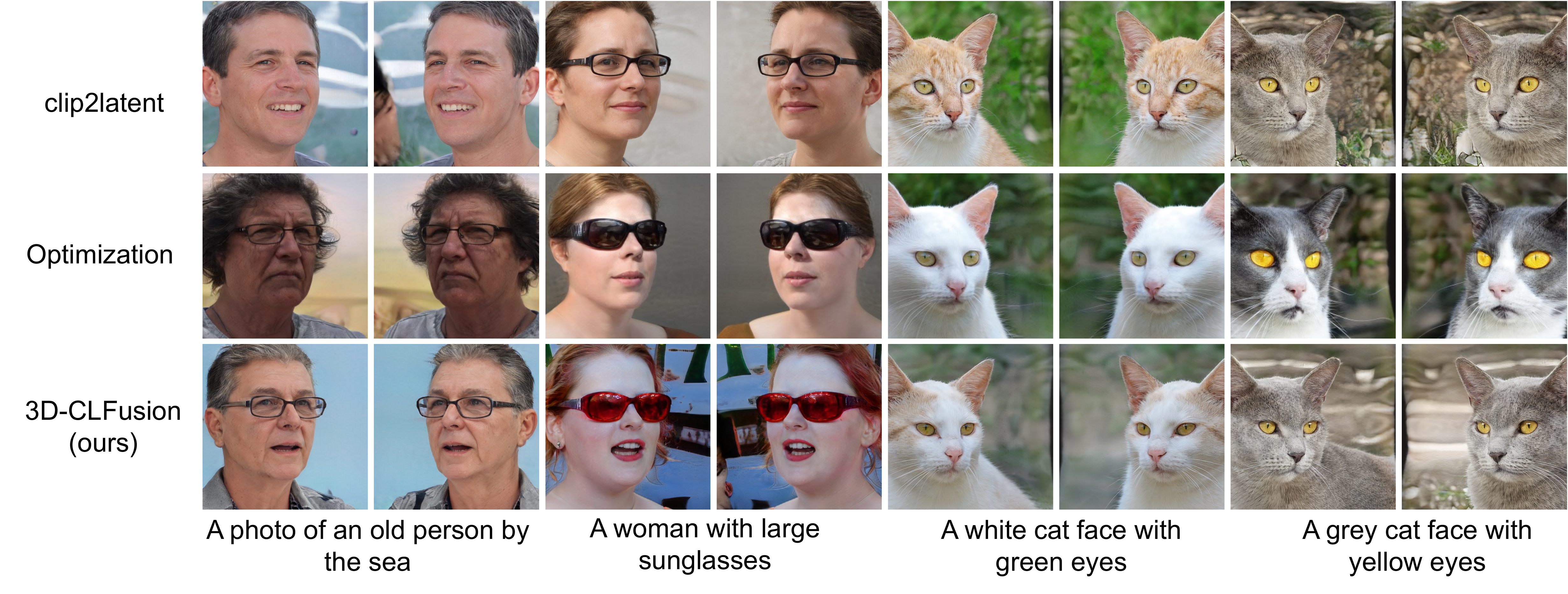}
  \vspace{-7mm}
  \caption{\textbf{Qualitative comparisons on text-to-3D with pre-trained latent-based NeRF: EG3D~\cite{chan2022efficient}.} All of the output images are rendered using the same camera poses and the checkpoints from FFHQ and AFHQ (Cat class) datasets.}
  \label{fig:EG3D}
\end{figure*}
\subsection{Datasets}

Since our pipeline relies on the pre-trained latent-based NeRF, we train the latent-based NeRF on some datasets including FFHQ~\cite{karras2019style}, AFHQ~\cite{choi2020starganv2}, and CompCar~\cite{yang2015large}.
\newline

\noindent\textbf{FFHQ}~\cite{karras2019style} id a face dataset which contains 70,000 face images. We assume the captured faces are mostly in the center. Though the size of the images is provided as 1024 $\times$ 1024 in the dataset, all of the images are resized into 256 $\times$ 256 for training the checkpoints.
\newline

\noindent\textbf{AFHQ}~\cite{choi2020starganv2}  is an animal face dataset that contains 15,000 high-quality images at 512×512 resolution and includes three categories of animals which are cats, dogs, and wildlife. Each category has about 5000 images. For each category, the dataset split around 500 images as a test set and provide all remaining images as a training set. Only the training images are used in the experiments.
\newline

\noindent\textbf{CompCars}~\cite{yang2015large} is a car dataset that contains 136,726 images capturing the different vehicles with various styles. The original dataset contains images with different aspect ratios. All of the images in the dataset are center-cropped and resized into 256 $\times$ 256.

\subsection{Experimental Settings}
\paragraph{Pre-trained latent-based NeRFs.}
To train our diffusion prior in our 3D-CLFusion, we use the pre-trained StyleNeRF~\cite{gu2021stylenerf} and EG3D~\cite{chan2022efficient} models as the generators trained on FFHQ, AFHQ, and CompCars. To have a fair comparison, we use the pre-trained models that are provided by original papers online. Specifically, StyleNeRF~\cite{gu2021stylenerf} provides checkpoints on FFHQ in 256, 512, and 1024 dimensions. EG3D~\cite{chan2022efficient} provides models trained on FFHQ and AFHQ (only cat category). Since StyleNeRF and EG3D do not provide the checkpoints for realistic cars such as the images in CompCars, we additionally train the checkpoint of the generator on cars using StyleNeRF for evaluation.

\paragraph{Baselines.}
Since our 3D-CLFusion is the first diffusion prior to leveraging latent-based NeRFs for text-to-3D, we compare it with the most similar baseline: clip2latent~\cite{pinkney2022clip2latent}. clip2latent is also a framework for $w$ latent diffusion while the main difference is their design model is for 2D StyleGAN and does not have the constraints on view-invariant learning for NeRFs. To further compare with the direct optimization method, we compare our model with latent vector optimization in $\mathcal{W}$~\cite{Karras2019stylegan2}.


\begin{table}[t!]
  
  \centering
  \caption{\textbf{Quantitative results on CLIP score and the inference time.} The CLIP score is measured only on the rendered frontal view for fair comparison. The results are measured on the StyleNeRF and Eg3D generators trained on FFHQ. The testing 64 prompts are from the ones provided by clip2latent~\cite{pinkney2022clip2latent}. The time is measured on one Nvidia 3090 GPU. The number in bold and underline indicate the best and the second-best results.}\label{table:abl_clip}
  \resizebox{\linewidth}{!}
  {
  \begin{tabular}{l|cc|c}
  \toprule
  %
  \multirow{2}{*}{Method}& \multicolumn{2}{c|}{CLIP score} &\multirow{2}{*}{Time} \\
  %
  \cmidrule{2-3} 
  & StyleNeRF~\cite{gu2021stylenerf} & EG3D~\cite{chan2022efficient} \\
  %
  %
  \midrule
  Ours (w/ $f_\theta$) & {0.337} & {0.291} &  18.4 s\\
  Optimization~\cite{Karras2019stylegan2} & {0.358} & {0.343} & 55.5 s\\
  \midrule
  DreamFusion~\cite{poole2022dreamfusion} & \multicolumn{2}{c|}{0.271} & 90 mins\\
  Magic3D~\cite{lin2022magic3d} & \multicolumn{2}{c|}{0.285} & 40 mins\\
  \midrule
  clip2latent~\cite{pinkney2022clip2latent} & 0.282 & 0.245 & 18.1 s\\
  Optimization~\cite{Karras2019stylegan2} & \textbf{0.358} & \textbf{0.343} & 55.5 s\\
  Ours (w/ $f_\theta$) & \underline{0.337} & \underline{0.291} &  18.4 s\\
  \midrule 
  Ours (w/ $\epsilon_\theta$) & 0.287 & 0.254 &  18.2 s\\
  Ours w/o $\mathcal{L}_2$ & 0.305 & 0.272 &  18.4 s\\
  Ours w/o $\mathcal{L}_{tri}$ & 0.311 & 0.282 &  18.4 s\\
  \bottomrule
  \end{tabular}
  }
  
\end{table}
\paragraph{Evaluation settings.}

\begin{figure*}[t!]
  \centering
  \includegraphics[width=0.95\linewidth]{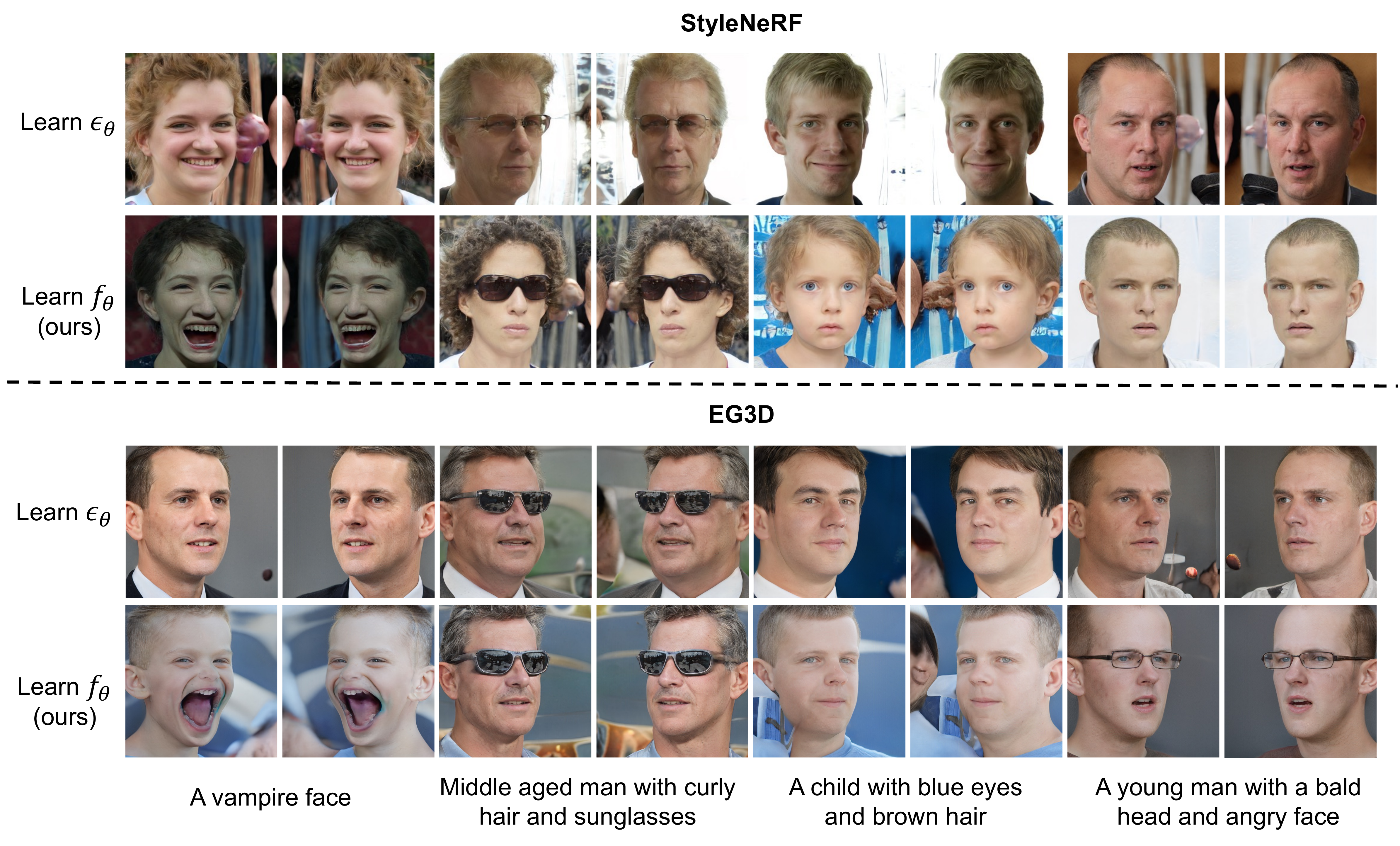}
  \caption{\textbf{Ablation studies on the learning of $\epsilon_\theta$ and learning of $f_\theta$ in the diffusion prior of 3D-CLFusion.} The results are produced from the generators of both StyleNeRF and EG3D, which are trained on the FFHQ dataset.}
  \label{fig:abl}
  \vspace{-4mm}
\end{figure*}

For qualitative comparison, we compare the input text prompt and the multi-view images generated from its latent code. We would like to note that, the generators trained on FFHQ and AFHQ are trained using the frontal head yaw angle roughly ranging between $-45^{\circ}$ to $45^{\circ}$ degrees for StyleNeRF and EG3D.
Only the checkpoints on CompCars can have 360-degree rendered images. For quantitative results, we use the CLIP similarity score following clip2latent~\cite{pinkney2022clip2latent} and only measure the frontal view of the latent-based NeRFs generated by the models trained on FFHQ.

\subsection{Implementation Details}

To train our diffusion priors (either $\epsilon_\theta$ or $f_\theta$), we leverage the generated images from the generators to generate the ground-truth paired data. Specifically, we sample a latent $w \in \mathbb{R}^{512}$ from the generator and generate $k=8$ views by manipulating camera poses for one $w$. Each generated image will be resized to $224 \times 224$ for the CLIP image encoder, \textit{i.e.,} ViT-B/32 we use in the experiments, to generate the CLIP embeddings $e \in \mathbb{R}^{512}$. To have the same comparison with clip2latent~\cite{pinkney2022clip2latent}, we also choose the same architecture as the diffusion prior in DALLE-2~\cite{ramesh2022hierarchical} where the $\epsilon_\theta$ or $f_\theta$ is implemented with causal transformer~\cite{melnychuk2022causal}. Following the training strategy in clip2latent, we also apply classifier-free guidance~\cite{ho2022classifier} and pseudo-text embeddings~\cite{zhou2021lafite} to enhance the diversity of the generated image and prevent overfitting on the image embeddings. The number of timesteps for the diffusion process is set as 1000. For the hyperparameter for all of the loss functions, all losses are equally weighted ($\lambda_{diff}=1.0$ and $\lambda_{contrast}=1.0$) for all the experiments. The batch size is set as $512$ where we ensure there are 64 ground-truth $w$ latent each with 8 different CLIP image embeddings from different views for optimizing the loss.  We optimize the network using Adam optimizer with a learning rate of $0.0001$ and train the model for 1,000,000 iterations. Each experiment is conducted on 1 Nvidia GPU 3090 (24G) and implemented in PyTorch.

\subsection{Results and Comparisons.}

In this section, we compare our proposed model with the baseline approach: clip2latent~\cite{pinkney2022clip2latent} and the online optimization method~\cite{Karras2019stylegan2} qualitatively (in Figure~\ref{fig:stylenerf} and Figure~\ref{fig:EG3D}) and quantitatively (in Table~\ref{table:abl_clip}). 
\paragraph{Qualitative results.}
As shown in Figure~\ref{fig:stylenerf}, we compare the quality of the generated latent $w$ among all of the models on the generated images from StyleNeRF. The left part displays the results on the FFHQ dataset while the right part displays the results on CompCars. There are some phenomena that can be summarized. First, although clip2latent is able to generate a latent code that can generate close images, semantically the output images are not well matched. For example, the gender on the FFHQ and the color of the vehicle on CompCars are not well matched with the input text prompts. This infers that without the constraints to ensuring the latent $w$ to be invariant, the generated latent code $w$ from the text CLIP embeddings may not share close latent $\mathcal{W}$ space. Second, the optimization is able to generate satisfactory results on FFHQ while failing to optimize the entire color of the vehicle. We credit the reason to that because the 360-degree vehicle is difficult to manipulate compared with faces in FFHQ. Thus, directly optimizing CLIP loss will be easy to bring artifacts and generate unnatural results (see the rightmost column of the blue SUV with green background).

We can also observe similar observations on EG3D~\cite{chan2022efficient} shown in Figure~\ref{fig:EG3D}. First, clip2latent fails to generate the well-matched latent for EG3D to render (see the sunglasses in the second big column and the color of the cat in the third big column). The optimization method is also able to generate well-matched results while they are unnatural (see the yellow eyes in the rightmost column). Our proposed methods exhibit comparable results as direct optimization.

\paragraph{Quantitative results.}

As shown in the first three rows of Table~\ref{table:abl_clip}, we compare the two methods with the CLIP similarity score using the 64 text prompts provided by clip2latent~\cite{pinkney2022clip2latent}. To measure the inference time and have a fair comparison, we train the optimization~\cite{Karras2019stylegan2} for 200 iterations for each text prompt. Our proposed model achieves a superior score compared with clip2latent with a similar inference time. On the other hand, although the optimization method has artifacts in qualitative results, it has the best result in CLIP score. This is basically because the objective of the optimization is sorely CLIP similarity. Thus, online optimization is supposed to have the best score. 

\begin{figure}[t!]
  \centering
  \includegraphics[width=\linewidth]{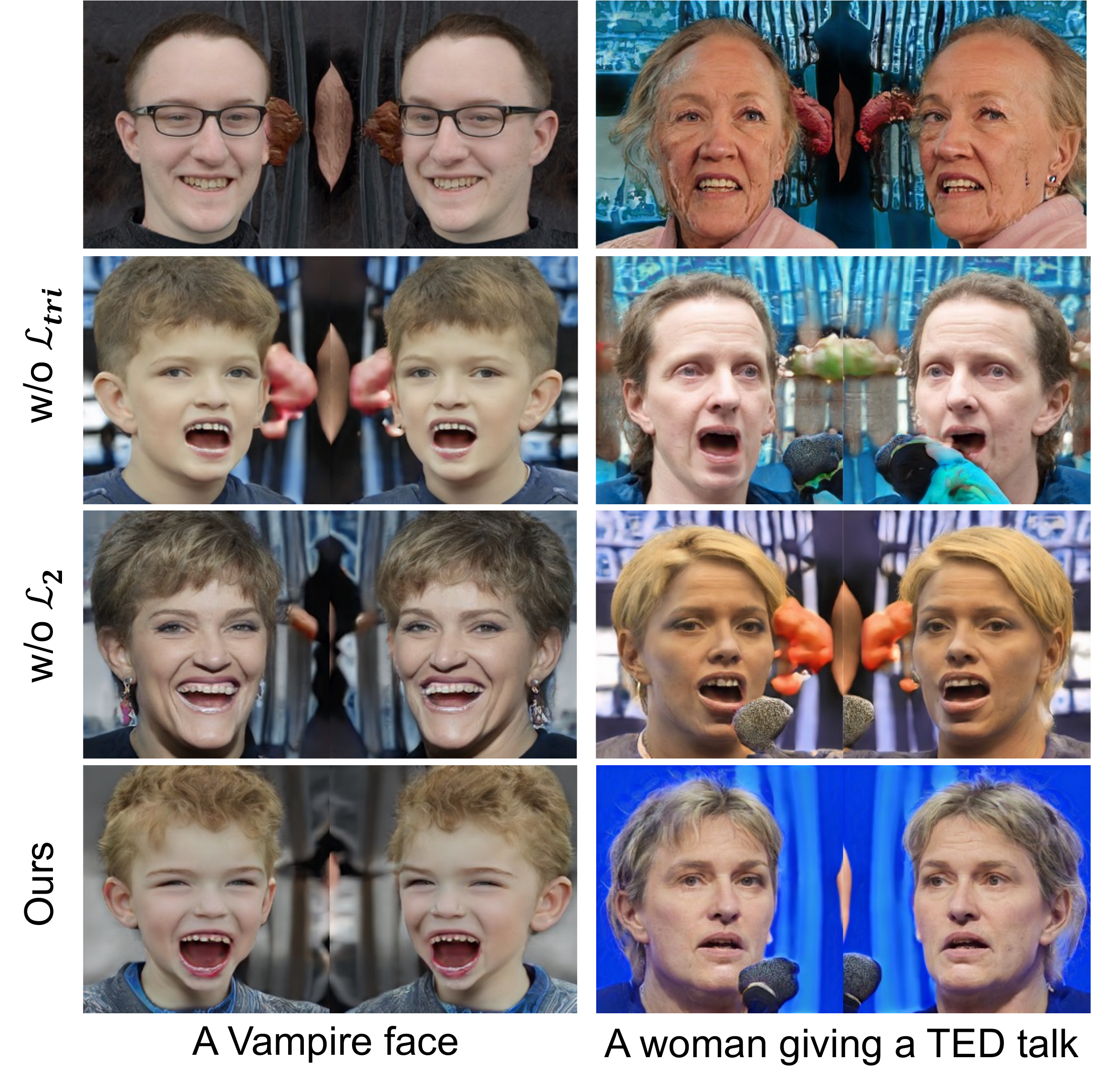}
  \caption{\textbf{Ablation studies on the proposed losses.} The results are from the StyleNeRF generator trained on FFHQ.}
  \label{fig:abl_loss}
  \vspace{-4mm}
\end{figure}

\subsection{Ablation studies}
\paragraph{Learning objectives: $\epsilon_\theta$ vs $f_\theta$.}
To further support the claim in Section~\ref{sec:method} that learning $f_\theta$ is more stable than $\epsilon_\theta$ when applying contrastive loss, we compare the models learning two different objectives qualitatively in Figure~\ref{fig:abl} and quantitatively in Table~\ref{table:abl_clip}. As shown in Figure~\ref{fig:abl}, it is obvious that learning $\epsilon_\theta$ would lead to inferior results. For example, learning $\epsilon_\theta$ can not generate good latent when feeding the input text prompt as "A vampire's face" in either StyleNeRF or EG3D. Table~\ref{table:abl_clip} also shows that learning $\epsilon_\theta$ exhibits a worse CLIP score compared with $f_\theta$. We credit the reason that by producing the $\Bar{w}_0$ from $\epsilon_\theta$, the $\Bar{w}_0$ will be far from consistency since it also depends on $w_t$ in each step. The view-invariant learning will not work easily on the network in $\epsilon_\theta$.

\paragraph{Contrative loss function: $\mathcal{L}_{contrast}$.}
To further analyze the effectiveness of essential contrastive losses of our proposed 3D-CLFusion, we conduct the experiments with one of them excluded and present the qualitative result in Figure~\ref{fig:abl_loss} and quantitatively in Table~\ref{table:abl_clip} as well. First, apparently without L2 loss $\mathcal{L}_{2}$, the diffusion model is not able to well derive view-invariant latent code in each diffusion process while it can still ensure the multi-view CLIP embeddings will produce close $w_0$ with the remaining triple loss. Second, we can observe that in the model with the triplet loss $\mathcal{L}_{tri}$ in our $\mathcal{L}_{contrast}$ excluded, the produced latent code $w$ starts to semantically deviate the input text prompt, which infers that the inter-class distance learning is also important for learning the view-invariant latent code in $\mathcal{W}$ latent space. Third, with the entire contrastive loss $\mathcal{L}_{contrast}$ excluded, the model is not able to have any guidance on view-invariant learning and would produce similar results as clip2latent~\cite{pinkney2022clip2latent}.

\section{Conclusion}

We have unveiled the challenges of the task text-to-3D directly using latent-based NeRFs and the limitations of the current models for this task. We propose a framework named 3D-CLFusion, which aims to produce view-invariant latent for rendering 3D objects with NeRFs from the input text prompt. Compared with the existing baseline models in the experiments, our designed model achieves more effective text-to-3D with pre-trained NeRFs. Though the 3D object created by our model is domain-specific to the pre-trained model, the inference time is at least 100 times faster than the existing text-to-3D creation from neural rendering. We would like to note that, our model will be applicable in the real world if more pre-trained NeRFs with vast categories of 3D objects are available.

{\small
\bibliographystyle{ieee_fullname}
\bibliography{egbib}
}
\clearpage
\appendix
\section{More results on multi-modal manipulation}

\begin{figure*}[t]
  \centering
  \includegraphics[width=0.7\linewidth]{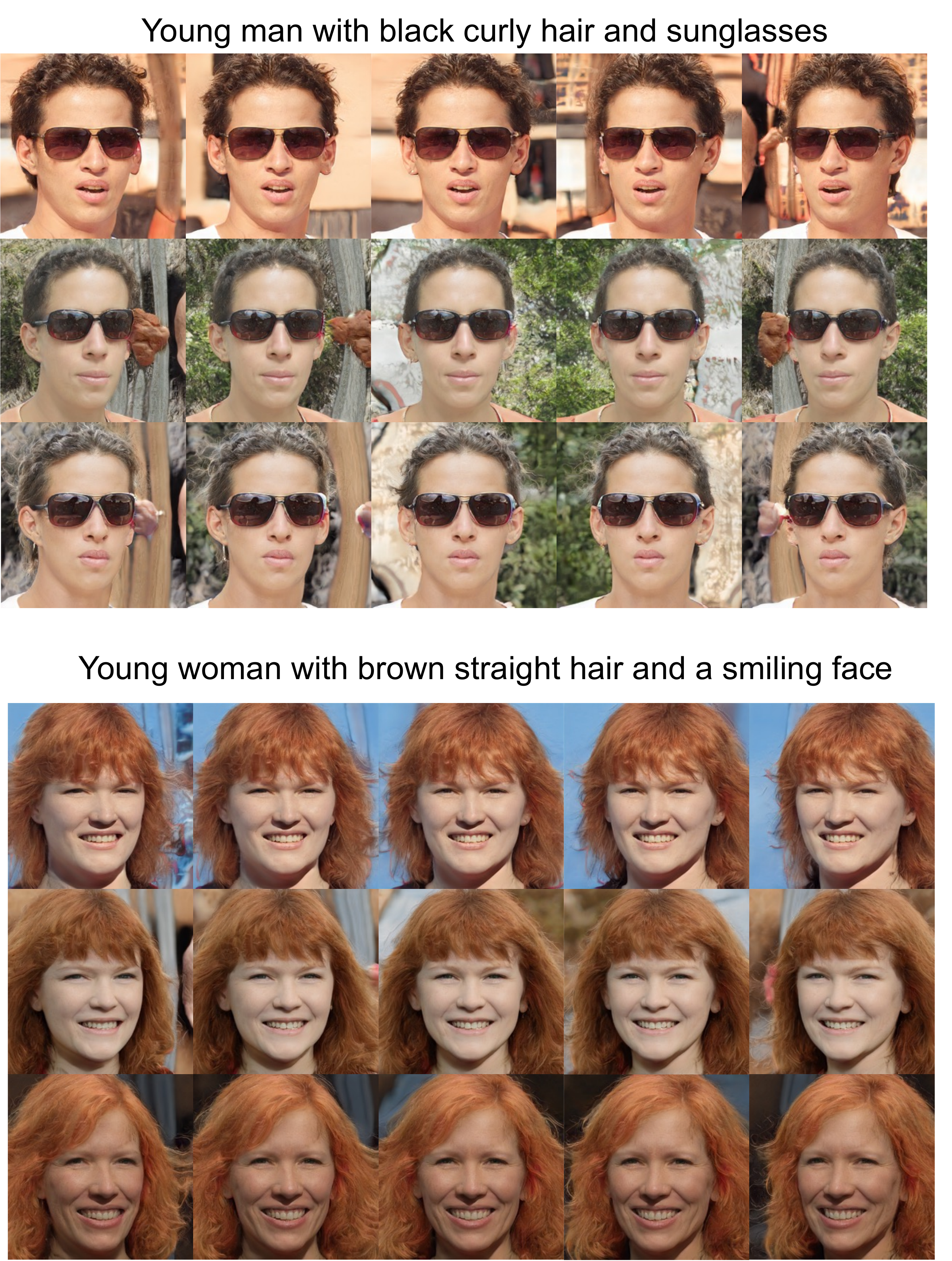}
  \caption{Results on StyleNeRF generator trained on FFHQ. Each row indicates the same input text with different input sampled noise.}
  \label{fig:morestylenerfffhq}
\end{figure*}

\begin{figure*}[t]
  \centering
  \includegraphics[width=0.7\linewidth]{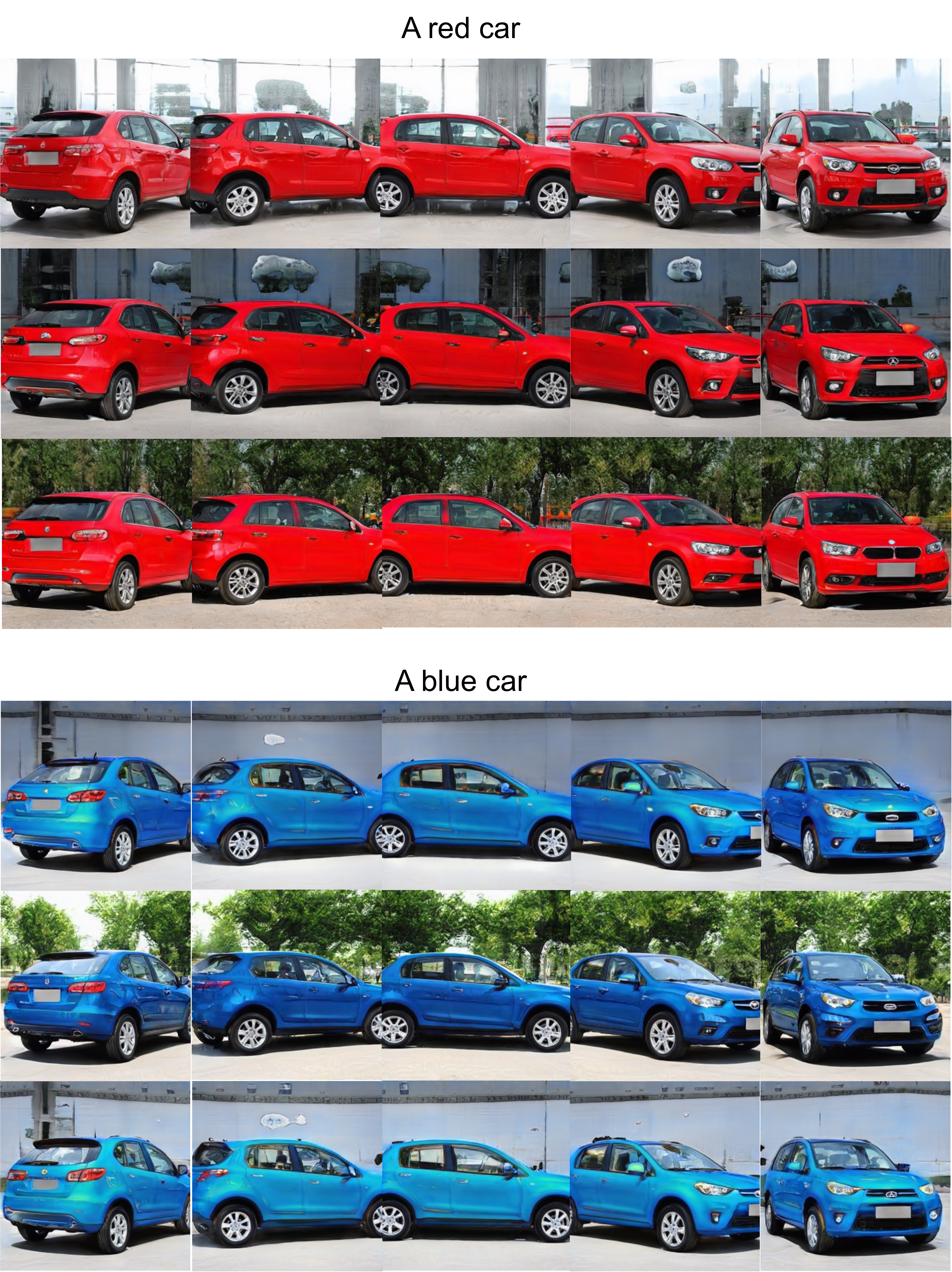}
  \caption{Results on StyleNeRF generator trained on CompCars. Each row indicates the same input text with different input sampled noise.}
  \label{fig:morestylenerfcar}
\end{figure*}

Since our 3d-CLFusion takes the input text embedding and the input gaussian noise as the starting latent for diffusion, we show examples of manipulating the random initial gaussian latent given the same input text prompt.
\paragraph{Generator employing StyleNeRF}
The results on FFHQ are presented in Fig~\ref{fig:morestylenerfffhq} and the results on CompCars are presented in Fig~\ref{fig:morestylenerfcar}.

\begin{figure*}[t]
  \centering
  \includegraphics[width=0.7\linewidth]{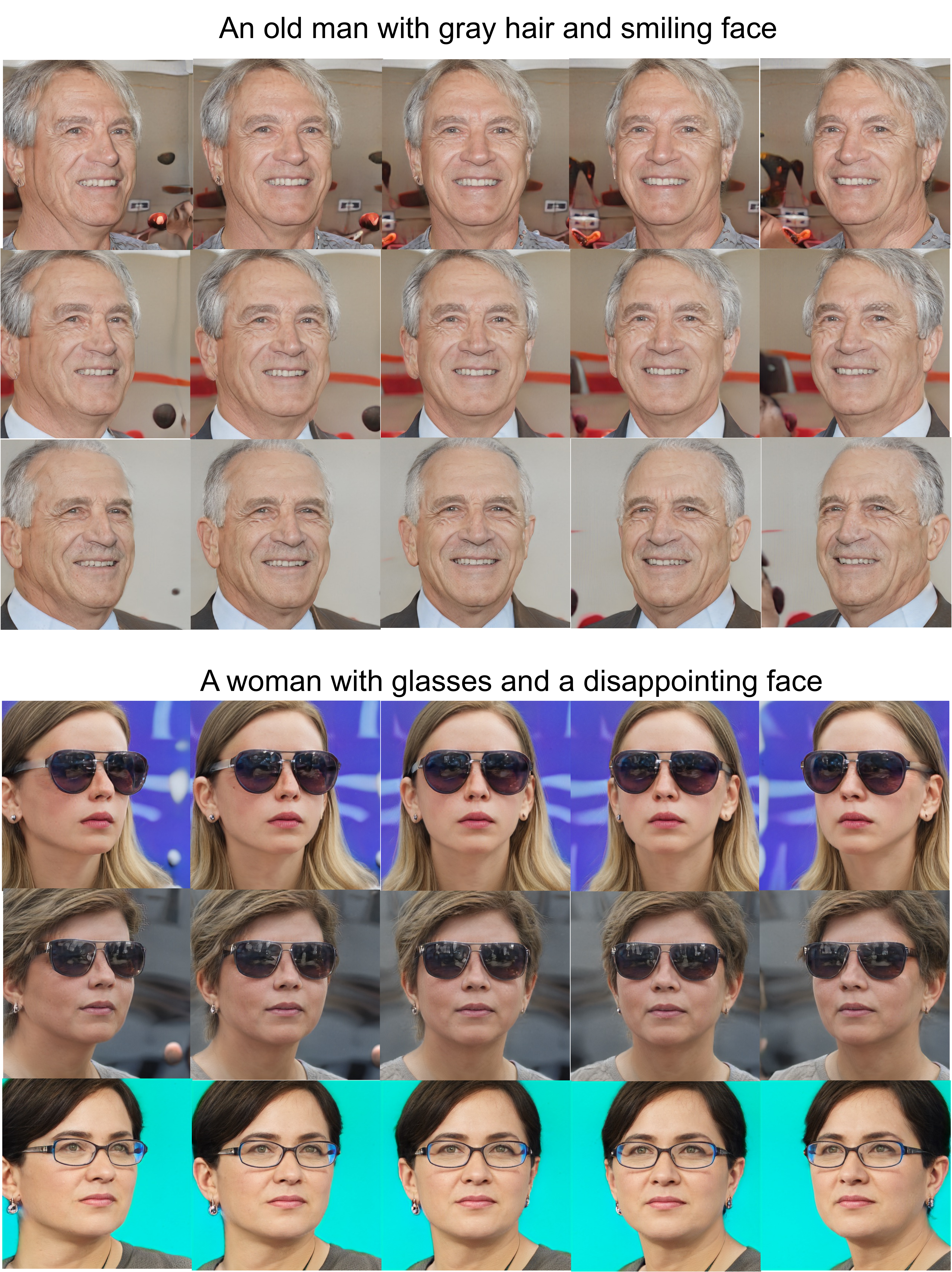}
  \caption{Results on Eg3D generator trained on FFHQ. Each row indicates the same input text with different input sampled noise.}
  \label{fig:moreeg3dffhq}
\end{figure*}

\begin{figure*}[t]
  \centering
  \includegraphics[width=0.7\linewidth]{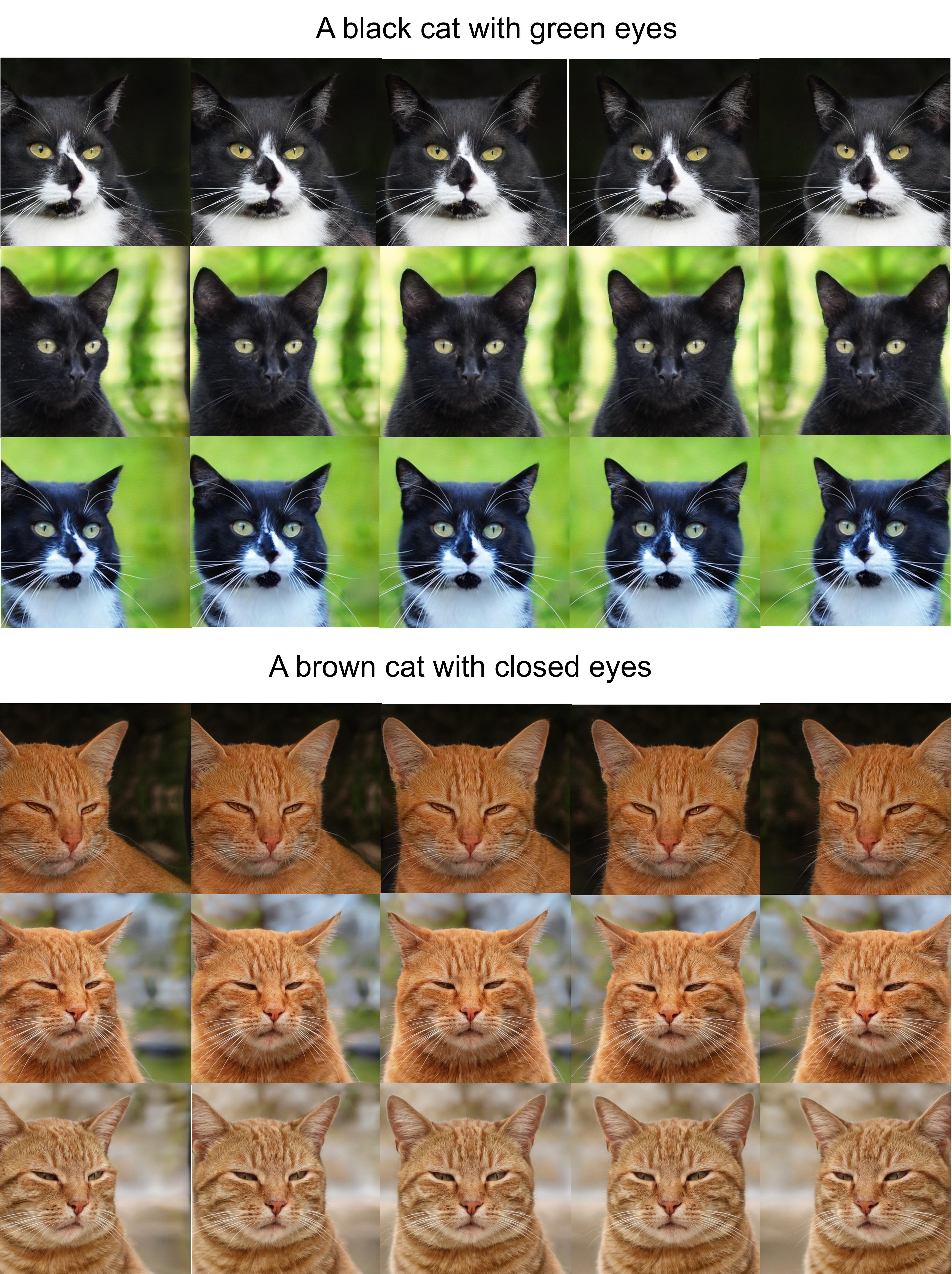}
  \caption{Results on Eg3D generator trained on AFHQ cats. Each row indicates the same input text with different input sampled noise.}
  \label{fig:moreeg3dcats}
\end{figure*}

\paragraph{Generator employing Eg3D}
The results on FFHQ are presented in Fig~\ref{fig:moreeg3dffhq} and the results on AFHQ cats are presented in Fig~\ref{fig:moreeg3dcats}.




\end{document}